\documentclass{article}
\PassOptionsToPackage{numbers,sort&compress}{natbib}

\usepackage[preprint]{neurips_2026}

\usepackage[utf8]{inputenc} % allow utf-8 input
\usepackage[T1]{fontenc}    % use 8-bit T1 fonts
\usepackage{hyperref}       % hyperlinks
\usepackage{url}            % simple URL typesetting
\usepackage{booktabs}       % professional-quality tables
\usepackage{amsfonts}       % blackboard math symbols
\usepackage{amsmath}        % align, cases, \text
\usepackage{amssymb}        % \triangleq, \varnothing
\usepackage{bm}             % \bm, bold math for Latin and Greek alike
\usepackage{graphicx}       % \includegraphics for the figure placeholders
\usepackage{nicefrac}       % compact symbols for 1/2, etc.
\usepackage{microtype}      % microtypography
\usepackage{xcolor}         % colors
\usepackage{comment}

\newcommand{\sg}{\operatorname{sg}}

\newcommand{\bfx}{\bm{x}}            % input / embedding
\newcommand{\bfz}{\bm{z}}            % block activation
\newcommand{\bftheta}{\bm{\theta}}   % block parameter vector
\title{ERASE: \emph{E}a\emph{R}ly b\emph{A}ckpropagation \emph{S}ch\emph{E}dule for Faster Training of Modern
Recommendation Systems}

\author{%
  Ergan Shang \\
  Carnegie Mellon University \\
  \texttt{eshang@andrew.cmu.edu} \\
  \And
  Flavio Sales Truzzi \\
  Meta Inc. \\
  \texttt{ftruzzi@meta.com} \\
}

\begin{document}

\maketitle

\begin{abstract}
Lightweight proxy models enable rapid experimentation without repeatedly
training frontier-scale systems, but their small kernels often leave modern
accelerators underutilized. Conventional training compounds this inefficiency
by scheduling the forward and backward passes as disjoint phases, so spare
capacity in one cannot be filled by work from the other. We reinterpret the
detachment mechanism of Forward-Forward (FF) as a scheduling primitive: given
a local objective, detaching a block's output removes downstream gradient
dependencies, making its backward pass ready when its forward pass finishes.
\textbf{\emph{ERASE}} launches each detached subgraph's backward pass early on a
separate CUDA stream, overlapping it with subsequent forward work. Execution
trace on a lightweight transformer demonstrates this overlap and its limit: a
kernel that saturates the device leaves no capacity for concurrency. On a
large-scale click-through-rate model, detaching six dense
subarchitectures improves training throughput by up to $9.51\%$ while keeping
normalized entropy close to the baseline.
\end{abstract}

\section{Introduction}
\label{sec:intro}

Backpropagation \citep{rumelhart1986learning,rumelhart19881986} remains the
standard method for training modern neural networks across vision and language
\citep{deng2009imagenet,vaswani2017attention}, agent learning and evaluation
\citep{phan2025humanity,li2026longhorizonterminalbenchtestinglimitsagents,zhou2026asibenchdawnartificialsuperintelligence},
political and social network inference
\citep{lyu2022understanding,shang2026inference}, and the natural sciences
\citep{roohani2024predicting,zhang2025genetic}. Among its most computationally
demanding applications are large-scale recommendation and ranking systems
\citep{cheng2016wide,covington2016deep}, which are continuously retrained on
data streams that can outpace a single training job
\citep{borisyuk2024lirank,kharitonov2019federated}. Training throughput
therefore directly affects model freshness and hardware resource burden
\citep{gupta2021training}.

For these workloads, throughput depends on accelerator utilization as well as
peak arithmetic performance \citep{zhu2025rankmixer}. Collective
communication, data movement, and scheduling dependencies can leave capacity
unused \citep{mudigere2022software,gupta2021training}. Conventional
reverse-mode differentiation traverses the graph in reverse topological order,
beginning the backward pass only after the entire forward pass has completed.
The two phases therefore occupy disjoint intervals, preventing either from
using spare capacity in the other.

The same utilization pressure appears at the opposite end of the scale:
frontier-scale models require too many resources for rapid, repeated experimentation,
so automated research relies on lightweight proxies such as NanoChat
\citep{karpathy2026autoresearch}, which retains a useful experimental signal at
orders-of-magnitude smaller scale. The accelerator, however, does not shrink
with the model. Small GEMMs may expose too little parallelism to approach peak
arithmetic throughput, leaving capacity unused even without launch gaps; CPU
scheduling computational overhead can reduce utilization further when gaps occur. We
therefore ask how to improve model FLOPs utilization (MFU) when the model is,
by design, too small to saturate the device.

Meanwhile, the Forward-Forward (FF) algorithm trains each block against a local
objective and passes a \emph{detached} activation to its successor
\citep{hinton2022forward}. This partitions the end-to-end graph into
block-local subgraphs: gradients do not cross block boundaries, and activations
need not be retained for a global backward pass. Importantly for scheduling, a
block's backward pass becomes ready once its forward pass and local objective
are complete, without waiting for the terminal loss or later blocks. It can
therefore overlap with subsequent forward passes, enabling work that the
conventional forward-then-backward schedule forgoes.

We turn this observation into \textbf{\emph{ERASE}}
(Section~\ref{sec:algorithm}), which dispatches each detached subgraph's
backward pass as soon as its forward pass returns. Experiments
(Section~\ref{sec:experiments}) demonstrate the intended overlap and its
saturation limit on NanoChat, then measure the throughput improvement on a
ranking model. Section~\ref{sec:caveats} analyzes the staggering introduced by
asynchronous dispatch.

\subsection{Related Work}
\label{sec:related}

\paragraph{Backpropagation-free learning.}
Backpropagation-free methods replace the global backward pass with block-local
targets. NoProp and DiffusionBlocks attach an auxiliary denoising variable to
each block \citep{li2026noprop,shing2026diffusionblocks}, borrowing their
training signal from diffusion models
\citep{ho2020denoising,shang2025predicting}; this permits detached activations
and block-local updates. Earlier work on Forward-Forward and layer-wise
learning likewise shows that local objectives can train competitive models
without gradients crossing block boundaries
\citep{hinton2022forward,lorberbom2024layer}.

\paragraph{CUDA Graphs and execution-level acceleration.}
CUDA Graphs capture and replay kernel sequences, reducing per-kernel CPU launch computational
overhead in small-kernel workloads
\citep{pytorch2021cudagraphs,ekelund2025boosting}; recent compiler work makes
capture more robust in PyTorch \citep{ghosh2025pygraph}. ERASE combines this
execution mechanism with FF-style detachment, which creates independent pieces
while preserving the usual loss and exact gradients within each piece. Early
backward launches each piece on a separate CUDA stream as soon as its forward
pass returns, while CUDA Graphs keep the asynchronous launch order consistent
across ranks.

\section{Backprop Free Algorithm}
\label{sec:algorithm}

\subsection{The Forward-Forward algorithm}
\label{sec:ff}

\begin{figure}[htbp]
\centering  \includegraphics[width=\linewidth]{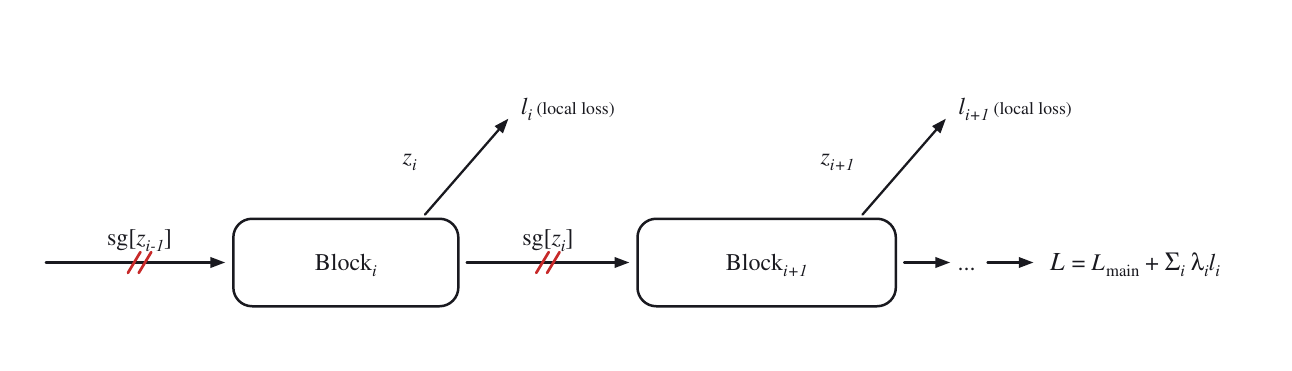}
  \caption{Schematic of FF-style detachment. Each block receives a detached copy
    of the previous block's activation, and contributes its own local loss. No
    gradient crosses a block boundary.}
  \label{fig:ff-detach}
\end{figure}

\paragraph{Data dependency shutdown.}
For an input embedding $\bfx \in \mathbb{R}^{p}$, consider a network of $B$
blocks with $\bfz_0=\bfx$, activations $\bfz_b$, and parameters $\bftheta_b$.
FF \citep{hinton2022forward} computes
\begin{equation}
  \label{eq:detach}
  \bfz_{b} \;=\; f_{b}\big(\sg[\bfz_{b-1}]; \bftheta_{b}\big),
  \qquad b = 1, \dots, B,
\end{equation}
where $\sg[\cdot]$ is the stop-gradient operator: it is the identity on the
forward pass and has zero Jacobian on the backward pass. A local loss on the
undetached $\bfz_b$ updates $\bftheta_b$ alone and may, for example, apply
binary cross-entropy to the goodness score
$G_b=\sum_j z_{bj}^2$. Consequently, $\nabla_{\bftheta_b}$ depends only on the
subgraph between adjacent cut points and is ready once the block's forward pass
and local loss are complete, before the terminal loss is evaluated. ERASE
exploits this scheduling consequence of detachment; Figure~\ref{fig:ff-detach}
illustrates the resulting graph.

\subsection{Early backward in \textbf{\emph{ERASE}}: scheduling, streams, and determinism}
\label{sec:early-bwd}
\label{sec:scheduling}
\label{sec:execution}

Detachment exposes independence but does not change the schedule: a single
backward call on the aggregate objective
\begin{equation}
  \label{eq:sumloss}
  \mathcal{L} \;=\; \mathcal{L}_{\text{main}} \;+\; \sum_{i} \lambda_{i} \,\ell_{i}
\end{equation}
still runs at the end of the step. ERASE instead launches each detached subgraph's
backward pass as soon as its forward pass returns, accumulating its parameter
gradients while leaving the undetached remainder to the usual end-of-step
backward. Because backward operations otherwise serialize with subsequent
forward work on the default CUDA stream, ERASE uses separate streams and events
to enforce the remaining dependencies. This allows the backward pass of block
$b$ to overlap with the forward pass of block $b+1$.

CPU dispatch introduces another tradeoff. Main-thread early backward blocks
further dispatch until its autograd call is enqueued. A
\texttt{ThreadPoolExecutor} avoids this stall but can vary the kernel and
collective order across ranks, creating stragglers that we call
\emph{staggering}. Capturing the affected subgraphs as CUDA Graphs fixes the
launch order. Section~\ref{sec:dpa-staggering} compares the blocking and
non-blocking variants.

\section{Experiments}
\label{sec:experiments}

As a small-scale sanity check via one A100 GPU, FF-style detachment on a $30$-layer MLP using MNIST dataset (\cite{deng2012mnist})
reduced compute-matched backward time by $58\%$ ($23.5$ to $9.9$\,ms) and total
batch time by $30\%$ ($41.2$ to $28.8$\,ms). We then evaluate
\textbf{\emph{ERASE}} on NanoChat \citep{karpathy2026autoresearch}, using one
A100 GPU to demonstrate overlap, and on a lightweight version of a large-scale
recommendation model, using eight H100 GPUs to measure throughput.

\subsection{NanoChat}
\label{sec:nanochat}

\begin{figure}[htbp]
  \centering
  \includegraphics{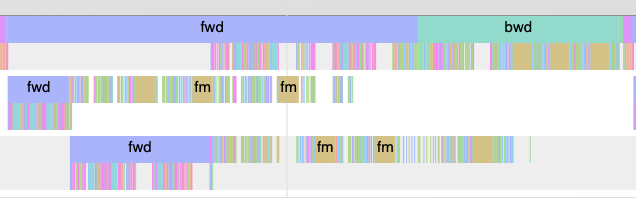}
  \caption{NanoChat single-batch trace with two detachment points and three
    streams. Backward work from one subgraph runs alongside forward work from
    the next.
  }
  \label{fig:nanochat-trace}
\end{figure}

To verify the intended overlap before experiments on a recommendation model, we partition
NanoChat's multi-head-attention stack into three subgraphs using two detachment
points. Figure~\ref{fig:nanochat-trace} shows a single-batch execution trace.

The trace demonstrates the execution pattern rather than measuring speedup:
one subgraph's backward work overlaps the next subgraph's forward work, as
Section~\ref{sec:early-bwd} predicts. The exception is the fused multi-head
attention backward kernel (\emph{fm} in Figure~\ref{fig:nanochat-trace}), which
runs alone. This is a resource limit rather than a dependency: by fusing the
attention matrix multiplications with softmax and avoiding materialization of
the sequence-by-sequence attention matrix, the kernel occupies every streaming
multiprocessor and leaves no capacity for concurrent work.

This limit reinforces the premise of Section~\ref{sec:intro}: overlap helps
only where the hardware is not already saturated. Reducing hidden width, depth,
or batch size shrinks kernels without shrinking the accelerator, leaving
resources that small kernels can share but device-filling kernels cannot.
Lightweight proxies contain mostly the former, making them natural targets for
early backward. Section~\ref{sec:dpa} quantifies the benefit on the recommendation
model.

\subsection{CTR Model}
\label{sec:dpa}

\subsubsection{Setup}
\label{sec:dpa-setup}

The remaining experiments use a click-through-rate
(CTR) model with six detached subarchitectures, four of which are assigned
separate CUDA streams alongside the main stream. Throughput is the number of
training examples processed per second (QPS), summarized by the post-warm-up
$p90$ of per-step samples (higher is better); unlike latency $p90$, this is the
fast end of the distribution. Quality is normalized entropy (NE), the model's
cross-entropy divided by that of a constant predictor; lower is better.

\subsubsection{Early backward on the CTR model}
\label{sec:dpa-early-bwd}

\begin{figure}[htbp]
  \centering
  \includegraphics[width=\linewidth, height=2cm]{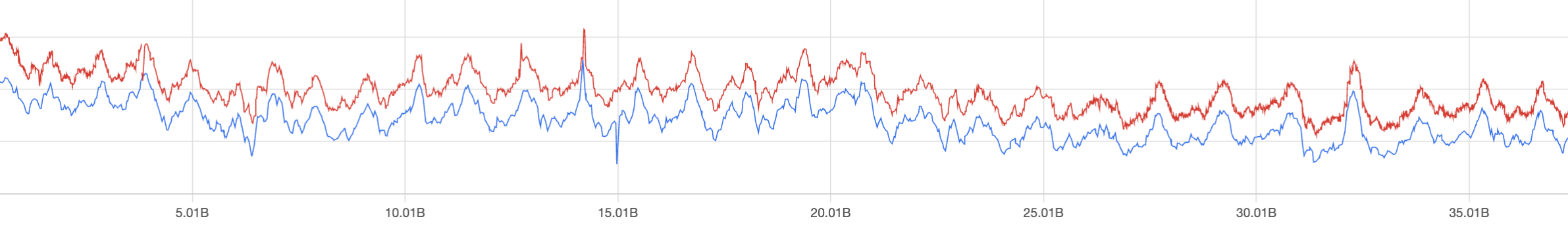}
  \caption{NE gap between the non-blocking early-backward variant with CUDA Graphs (red curve) and the baseline (blue curve),
    approximately $1.38\%$. }
  \label{fig:ne-gap}
\end{figure}

ERASE launches each detached subarchitecture's backward pass from a
\texttt{ThreadPoolExecutor} as soon as its forward pass completes. Because the
worker thread can introduce cross-rank launch-order nondeterminism, we capture
a subset of the subarchitectures as CUDA Graphs to fix their order and reduce
CPU dispatch computational overhead. This non-blocking configuration (row~1 of
Table~\ref{tab:qps}) improves throughput by $7.38\%$ with an NE gap of
approximately $1.38\%$ (Figure~\ref{fig:ne-gap}).
Section~\ref{sec:caveats} examines its FUP=False setting and compares it with
blocking dispatch.

\section{Caveats}
\label{sec:caveats}
\label{sec:dpa-staggering}

Table~\ref{tab:qps} reports dispatch and FUP
(\texttt{find\_unused\_parameters}) ablations. \emph{Blocking} launches early
backward from the main thread and stalls CPU dispatch during autograd enqueue;
FUP controls which parameters participate in gradient synchronization.

\begin{table}[htbp]
  \caption{Throughput on the CTR model, $p90$ queries per second
  }
  \label{tab:qps}
  \centering
  \begin{tabular}{llrr}
    \toprule
    \# & Configuration & QPS ($p90$) & Gain vs.\ baseline \\
    \midrule
    -- & baseline                  & 184{,}535.51 & $+0.00\%$ \\
    3 & blocking, FUP=False       & 185{,}223.71 & $+0.37\%$ \\
    2 & blocking, FUP=True & 194{,}244.72 & $+5.26\%$ \\
    1  & non-blocking + CUDA Graphs, FUP=False  & 198{,}145.28 & $+7.38\%$ \\
    0  & blocking, FUP=True$^{\dagger}$  & \textbf{202{,}078.43} & $\mathbf{+9.51\%}$ \\
    \bottomrule
  \end{tabular}
\end{table}

Blocking is deterministic but stalls the CPU: FUP=False gains only $0.37\%$
(row~3), whereas FUP=True gains $5.26\%$ (row~2) by reducing the
parameters in the final aggregate backward. CUDA Graphs instead recover a
$7.38\%$ gain with non-blocking dispatch and FUP=False (row~1). Together, these
results suggest that deterministic, rank-synchronized collective order is
useful and that either FUP or CUDA Graphs can provide it.

\section{Conclusion}
\label{sec:conclusion}

\textbf{\emph{ERASE}} repurposes FF-style detachment as a scheduling primitive:
cutting inter-block dependencies makes each backward pass ready after its
forward pass, enabling separate-stream overlap while preserving exact
within-subgraph gradients without FF's goodness objective. On a CTR
model, ERASE improves $p90$ QPS by $5$--$9\%$ with a small NE gap under
deterministic cross-rank collective order.

\medskip

\bibliographystyle{plainnat}
\bibliography{references}

\end{document}